\documentclass[lettersize,journal]{IEEEtran}
\usepackage{flushend}
\usepackage{amsfonts}
\usepackage{textcomp}
\usepackage{xcolor}
\usepackage{fancyhdr}
\usepackage{cite}
\usepackage{graphicx}
\usepackage{psfrag}
\usepackage{subfigure}
\usepackage{url}
\usepackage{stfloats}
\usepackage{amsmath}
\usepackage{array}
\usepackage{fancyhdr}
\usepackage{epsfig}
\usepackage{amssymb}
\usepackage{color}
\usepackage{float}
\usepackage{algorithm}
\usepackage{algpseudocode}
\usepackage{balance}
\usepackage{soul}
\usepackage{caption}
\usepackage{amsthm}
\usepackage{balance}
\usepackage[caption=false,font=normalsize,labelfont=sf,textfont=sf]{subfig}

\def\BibTeX{{\rm B\kern-.05em{\sc i\kern-.025em b}\kern-.08em
    T\kern-.1667em\lower.7ex\hbox{E}\kern-.125emX}}
\begin{document}
\title{On-Air Deep Learning Integrated Semantic Inference Models for Enhanced Earth Observation Satellite Networks: A Review Beyond Horizons}
\author{\IEEEauthorblockN{Hong-fu Chou,~\IEEEmembership{Member,~IEEE}, Vu Nguyen Ha,~\IEEEmembership{Senior Member,~IEEE},  Prabhu K. Thiruvasagam,~\IEEEmembership{Member,~IEEE}, 
Thanh-Dung Le,~\IEEEmembership{Member,~IEEE}, Geoffrey Eappen,~\IEEEmembership{Member,~IEEE}, Ti Ti Nguyen,~\IEEEmembership{Member,~IEEE}, Duc Dung Tran,~\IEEEmembership{Member,~IEEE}, 
Luis M. Garces-Socarras,~\IEEEmembership{Member,~IEEE}, Jorge L. Gonzalez-Rios,~\IEEEmembership{Member,~IEEE}, Juan Carlos Merlano-Duncan,~\IEEEmembership{Senior Member,~IEEE}, and Symeon Chatzinotas,~\IEEEmembership{Fellow,~IEEE}}\thanks{This work was funded by the Luxembourg National Research Fund (FNR), with the granted SENTRY project corresponding to grant reference C23/IS/18073708/SENTRY.}
\thanks{ Hong-fu Chou, Vu Nguyen Ha,  Prabhu Kaliyammal Thiruvasagam, Thanh-Dung Le, Geoffrey Eappen, Ti Ti Nguyen, Duc Dung Tran, Luis M. Garces-Socarras, Jorge L. Gonzalez-Rios, Juan Carlos Merlano-Duncan, Symeon Chatzinotas are with the Interdisciplinary Centre for Security, Reliability and Trust (SnT), University of Luxembourg, Luxembourg (E-mail of  corresponding author: hungpu.chou@uni.lu).}}
\markboth{IEEE Journal of  Selected Topics in Applied Earth Observations and Remote Sensing,~Vol.~XX, No.~X, XX~202X}%
{Hong-fu Chou~\textit{et al.}:}
\maketitle

\begin{abstract}
Earth Observation (EO) systems are crucial for cartography, disaster surveillance, and resource administration. Nonetheless, they encounter considerable obstacles in the processing and transmission of extensive data, especially in specialized domains such as precision agriculture and real-time disaster response. Earth observation satellites, outfitted with remote sensing technology, gather data from onboard sensors and IoT-enabled terrestrial objects, delivering important information remotely. Domain-adapted Large Language Models (LLMs) provide a solution by enabling the integration of raw and processed EO data. Through domain adaptation, LLMs improve the assimilation and analysis of many data sources, tackling the intricacies of specialized datasets in agriculture and disaster response. This data synthesis, directed by LLMs, enhances the precision and pertinence of conveyed information. This study provides a thorough examination of using semantic inference and deep learning for sophisticated EO systems. It presents an innovative architecture for semantic communication in EO satellite networks, designed to improve data transmission efficiency using semantic processing methodologies. Recent advancements in onboard processing technologies enable dependable, adaptable, and energy-efficient data management in orbit. These improvements guarantee reliable performance in adverse space circumstances using radiation-hardened and reconfigurable technology. Collectively, these advancements enable next-generation satellite missions with improved processing capabilities, crucial for operational flexibility and real-time decision-making in 6G satellite communication.
\end{abstract}
\section{Introduction}
Earth Observation (EO) systems collect and analyze global data for many civic applications and essential services that profoundly affect human life via satellite networks. The duties include cartography, meteorological forecasting, risk assessment, disaster surveillance, natural resource administration, and emergency response. The projected value of the satellite-based Earth Observation market is anticipated to attain 11.3 billion USD by 2031 \cite{denis2017towards}. The Internet of Things (IoT) has resulted in substantial progress in information and communication technologies, hence augmenting the need for lucrative value-added services and applications. The proliferation of extensive Earth Observation data has been enabled by advancements in sensor, transmission, and storage technologies. This EOdataset has substantial information, although it poses challenges due to its unique attributes, including discrete observations and complex dimensions. The analysis of image retrieval methodologies seeks to tackle the difficulties presented by complex data \cite{li2021image}. The research highlights several challenges, including the limited availability and variety of training data, as well as the incapacity of existing algorithms to replicate human cognitive functions. 

Although there is a large amount of EO data available, there is a lack of effective conversion into actionable information for specialized sectors such as agriculture or disaster response. Big data systems that have frequent satellite revisits have the potential to revolutionize real-time monitoring \cite{li2020real}. NewSpace constellations \cite{koechel2018new} are expanding their capabilities beyond the provision of affordable and frequent data. They focus on addressing particular requirements by using specialized satellites and providing data analysis in addition to raw data, serving a broader spectrum of consumers. Furthermore, monitoring strategies need careful consideration of both geographical and temporal dimensions \cite{acevedo2023real}. The extent of the phenomena being examined might vary from a single spot to an intricate three-dimensional grid. Resolution, which refers to the level of detail in measurements, and expanse, which indicates the region covered, are both important aspects. An example of researching the influence of topography on soil moisture may include the measurement of moisture levels at many locations along a slope, necessitating a high level of spatial resolution. Consequently, current EO systems are facing difficulties in managing the substantial volume of data being sent and fulfilling the diverse demands for the latency at which the data has to be processed.  In order to address these significant challenges, the primary strategies involve two actions: firstly, improving the bandwidth of satellite communication networks that facilitate EO systems, and secondly, reducing data volume by employing on-board processors on the satellites to preprocess the EO data.

Utilizing the inter-satellite links (ISLs) of the multi-satellite, multi-orbit connectivity systems has been seen as a potential strategy, as stated in the first direction \cite{al2022survey}. The integration of several satellites and orbits, spearheaded by non-geostationary satellites (NGSO), is on the verge of transforming worldwide communication. Unlike conventional geostationary satellites (GSO), these systems provide reduced latency and loss as a result of their lower orbits. This research presented in \cite{chougrani2023connecting} examines the communication in multi-satellite, multi-orbit connectivity systems, with a special emphasis on NGSO technology. It investigates the possibilities of this technology while recognizing the obstacles associated with its adoption. The data acquired by EO satellites may undergo processing, storage, and transmission to Earth using SatCom satellites positioned in various orbits.  By using the ISLs, several routing channels may be constructed between the remote-sensing satellites and their respective ground stations. The technique of \cite{radhakrishnan2016survey} may effectively handle the problem of overload and backhaul bottlenecks that occur at certain nodes in inter-satellite communication (ISC) networks, resulting in improved network capacity.

The transmission of the meaning behind data is the primary objective of semantic communication, rather than the precise data itself. This has the potential to enhance efficiency by providing greater flexibility in encoding and decoding. The authors of \cite{Paolo2023} suggest a method that is predicated on three concepts: the use of a specialized data format to capture meaning, the adaptation of transmission based on channel conditions, and the use of generative models to adjust transmission and enable tasks at the receiver. Researchers of \cite{zhu2024flood} have devised a system that utilizes natural language conversations to disseminate flood risk information, with the aim of enhancing public preparation for floods. This system utilizes a large language model (LLM) that has been trained specifically on flood-related knowledge, guaranteeing the precision and reliability of the information it provides. Through the use of geographic information systems (GIS), the LLM has the ability to tailor risk evaluations according to the geographical location of the user. Utilizing natural language conversations has shown the potential to enhance flood risk perception among individuals with diverse cognitive capacities. Furthermore, semantic Importance-Aware Communication (SIAC) uses LLM to understand the meaning and semantics of conveyed data \cite{guo2023semantic}. This understanding allows the system to give priority to significant information by allocating more resources to data with greater semantic significance. Unlike earlier approaches, SIAC can seamlessly integrate into current communication networks with minimal alterations. By prioritizing crucial information, SIAC reduces semantic loss more effectively than conventional communication techniques.

In relation to the second aspect, the use of semantic communication (SemCom) principles has been recognized as a favorable strategy. The primary purpose of SemCom is to prioritize semantic accuracy above reducing bit/symbol error rates at the receiver side, which is in complete alignment with the core goals of most EO methods. The first principle of SemCom is the extraction of semantic data, wherein pertinent information is initially obtained, condensed, or broken down from unprocessed data. Afterward, just the information related to meaning is transferred to the central core. The PhiSat-1 project, presented in \cite{giuffrida2021varphi} by the European Space Agency (ESA), was among the first demonstrations of this concept. With the recent launch of PhiSat-2, this approach is now advancing to new capabilities. Two 6U CubeSats are deployed in this mission to capture Earth photographs. The data is processed inside the CubeSats to filter out information depending on the cloud coverage of the scene. The second concept revolves around semantic-oriented network management, specifically SemCom. SemCom's objective is to oversee all network functionalities across various layers (PHY, MAC, Network, Transportation, Application) by taking into account the semantics of the data and the requirements of applications related to EO tasks \cite{wang2016earth}.

\begin{figure*}[htbp]
\centerline{\includegraphics[width=160mm]{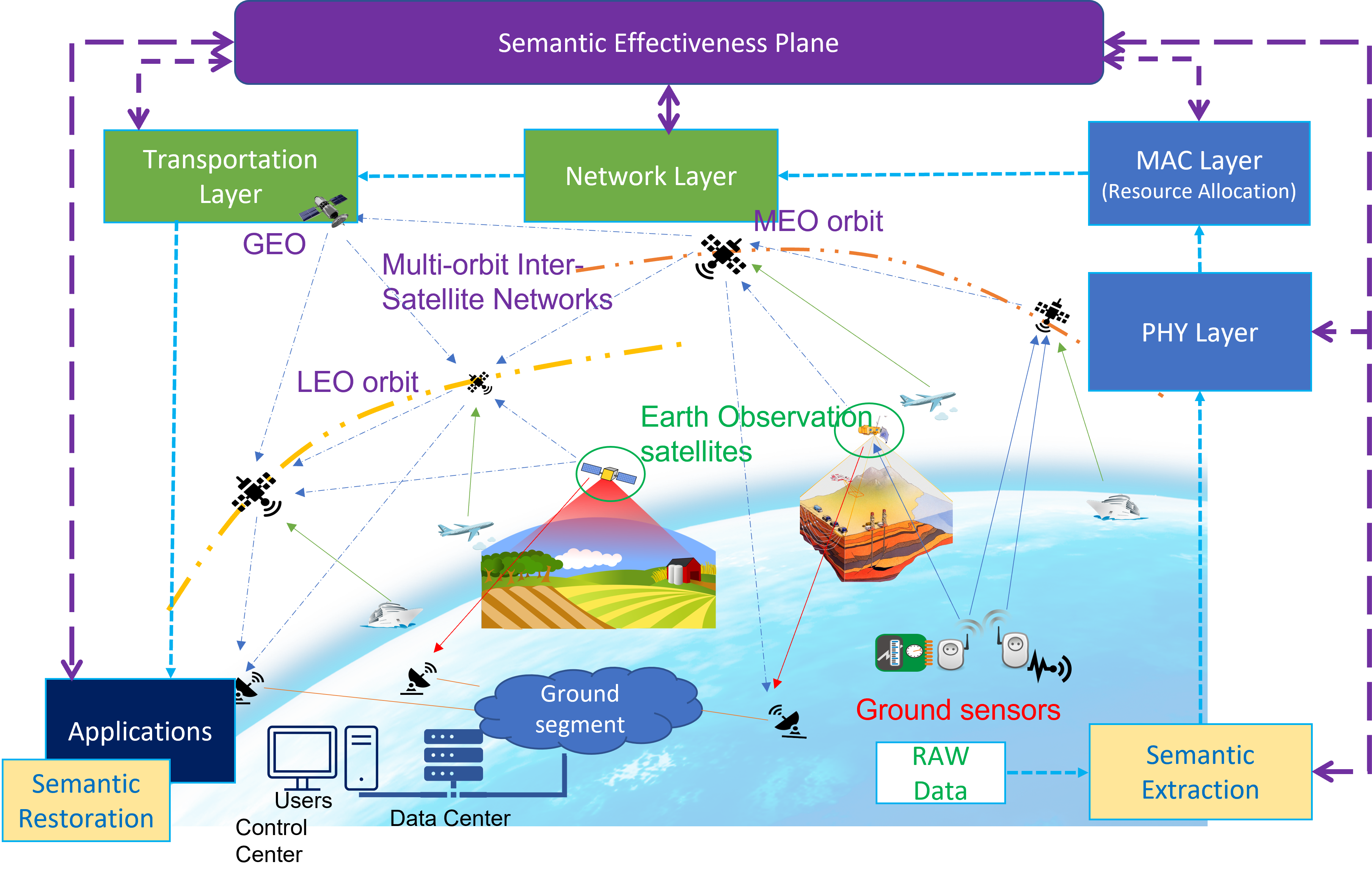}}
\caption{The system model of semantic-oriented EO satellite networks.}
\label{SystemModel}
\end{figure*}

Our approach introduces a systematic method for optimizing data flow and processing across multi-orbit EO networks, offering solutions to the major challenges of managing large-scale, diverse EO datasets. The purpose of presenting Fig. \ref{SystemModel} is to visually illustrate the architecture of the proposed multi-orbit inter-satellite network and to clarify the layered structure of satellite communication and data processing. This architecture helps contextualize our approach by demonstrating how Earth observation satellites in various orbits (LEO, MEO, GEO) can coordinate to gather and integrate data from ground sensors in diverse environments, such as agricultural fields and maritime regions. By providing this visual model, we aim to make the complex network of multi-orbit data flow and processing more accessible, highlighting the systematic approach we propose for optimizing data integration within Earth observation systems. The EO data, initially processed for semantic content from its raw form, is then transmitted to a "Ground segment", which includes data centers and user control interfaces for further analysis and user access. Data travels through multiple layers of communication within the network. At the base level, the PHY Layer (Physical Layer) ensures the fundamental transmission of data, followed by the MAC Layer (Medium Access Control) responsible for resource allocation. Moving upwards, the Network Layer manages the interconnectivity across the satellite network. The Transportation Layer handles the movement of data across the system, connecting the ground segments to various orbital satellites.

Above the operational layers is the Semantic Effectiveness Plane, which oversees and analyzes the flow of semantic data. The model features two key processes: Semantic Extraction and Semantic Restoration. At the ground level, data extracted from sensors undergoes semantic processing—either through ground-based AI servers or onboard edge computing—where meaningful patterns and insights are identified. This processed semantic information is subsequently restored into actionable applications, ranging from environmental monitoring to disaster management. The model proposes a closed-loop system in which raw data from ground sensors is transformed into semantically rich insights, facilitating seamless communication between satellite networks in space and ground infrastructure.

We summarize our contributions as follows.
\begin{itemize}
    \item We present an in-depth review of the latest developments in deep learning for semantic data processing in EO systems and 6G communication, emphasizing on-board processing and integration for segmentation, data compression, and pixel-level categorization to improve remote sensing analysis and retrieval. Additionally, we examine the evolution of SemCom, highlighting efficiency improvements through joint source-channel coding (JSCC) and deep learning, while exploring frameworks that enhance LLM adaptability and decision-making in networking and EO using task-specific models and advanced data fusion.
    \item The cognitive semantic EO system model enhances satellite networks by leveraging  Discrete Task-Oriented JSCC (DT-JSCC) \cite{xie2023robust} and Semantic Data Augmentation (SA) \cite{pu2024fine} techniques to transmit task-relevant semantic features rather than raw data, significantly reducing communication overhead and energy consumption in 6G satellite communication.
    \item The proposed framework incorporates inter-satellite communication and deep learning models for on-board processing, optimizing the transmission, analysis, and decision-making of multi-spectral EO data through an end-to-end semantic-augmented communication process, leading to improved classification accuracy and system efficiency.
\end{itemize}
The rest of the paper is structured as follows. Section II discusses semantic data processing from semantic data compression and fusion points of view to enhance the performance of applications that are based on EO data. Section III presents the evolution of SemCom networks, their attributes, and source and channel coding aspects with a framework for semantic inference-based EO. Section IV discusses the role of LLMs in enhancing domain-specific semantic effectiveness and automating EO. Section V introduces the SemCom-based, resource-efficient EO system, showcasing advancements in on-board processing hardware—such as SRAM-based FPGAs, MPSoCs, and reconfigurable designs—that pave the way for adaptive, resilient, and energy-efficient satellite systems capable of meeting the dynamic demands of modern space missions. Section VI concludes the work with a future outlook.
\section{Semantic data processing for EO applications}
This section delves into the concept of semantic segmentation, a technique that assigns precise identifiers to individual pixels in photographs. It concentrates on the application of this methodology to the examination of satellite imagery and other types of EO data. The content explores the application of semantic segmentation in a variety of applications, including trajectory facilitation, information retrieval, and picture compression. We also underscore the current advancements in deep learning for the classification of various EO data types using semantic knowledge. Irrespective of the intended meaning of the source information, traditional communication, based on the Shannon theory, emphasizes the reliable reception of individual sent bits. In contrast, the SemCom focuses on the successful transmission of only necessary semantic information to convey the meaning of the source message and thus it aspires to achieve the information capacity beyond the traditional Shannon paradigm \cite{strinati20216g}. Semantic-information feature learning \cite{Xu2022} is the key to addressing the utilization of cognitive techniques employed as a means to direct computational resources towards activities of higher importance. The process of achieving dynamic adaptation in semantic compression involves the utilization of a feature learning module and an attention feature module. These modules enable the source encoder to generate a variable number of symbols and modify the capabilities of the source encoder and channel encoder through the use of cognitive techniques.
The evolution of semantic communication \cite{Lu2022SemanticsEmpoweredCA} is traced back to the early 20\textit{th} century with its continuous growth into the realm of modern communications beyond 5G and 6G technologies. In light of this, there is a significant need to develop more intelligent and efficient communication protocols that can meet the diverse quality of service (QoS) needs. This must be done while addressing the challenge of limited communication bandwidth and computation. The development of an intelligent communication system is considered essential in both industry and academia. Such a system is not limited to memorizing data flows based on rigorous regulations but also aims to comprehend, analyze, and articulate the fundamental semantics.  In order to facilitate the implementation of immediate intelligent services in future communication systems, it is advantageous to distill and communicate only the information that is pertinent to the task at hand and precise semantics. The semantic approach effectively reduces the overall latency of the system.

Achieving accurate categorization at the pixel level is crucial in applications like driverless cars, where the capacity to understand lane markings and precisely identify the position of objects is essential. However, the process of semantic segmentation may be complicated by barriers such as hidden objects or congested backgrounds. Notwithstanding these challenges, semantic segmentation offers benefits that go beyond autonomous vehicles. The technology enables the examination of medical images by identifying abnormalities, assists robots in navigating their surroundings, and improves video surveillance by detecting potentially troublesome behavior. Recent progress in deep learning has led to remarkable achievements in achieving high precision at the pixel level \cite{younesi2024comprehensive}. However, its performance is inferior to that of normal satellite images when dealing with non-standard data such as point clouds and spectrum imaging. Likewise, a significant disparity in performance occurs when dealing with limited data sets. The lack of comprehensively classified, unconventional remote sensing data hinders the development and evaluation of innovative deep-learning methods.
\begin{table*}[h]
\centering
\caption{Summary of Methodologies and Applications in Semantic Image Processing and Compression}
\scriptsize
\begin{tabular}{|c|c|c|c|c|c|c|c|c|c|c|c|c|c|c|c|c|}
\hline
\textbf{Feature} & \textbf{\cite{jia2013feature}} & \textbf{\cite{espinoza2013earth}} & \textbf{\cite{augustin2019semantic}} & \textbf{\cite{ivashkovych2022corsa}} & \textbf{\cite{tran2020semantic}} & \textbf{\cite{li2021image}} & \textbf{\cite{xia2017exploiting}} & \textbf{\cite{yan2013semantic}} & \textbf{\cite{gao2019semantic}} & \textbf{\cite{richter2012semantic}} & \textbf{\cite{han2017compress}} & \textbf{\cite{liu2021semantics}} & \textbf{\cite{hong2023cross}} & \textbf{\cite{adam2023deep}} & \textbf{\cite{wang2024samrs}} & Proposed\\ \hline
EO data processing & \checkmark & \checkmark & \checkmark & \checkmark & \checkmark & & & & & \checkmark & \checkmark & & \checkmark & &\checkmark &\checkmark\\ \hline
AI-based compression & & & & \checkmark & & \checkmark & & & & & & & \checkmark & & &\checkmark\\ \hline
Semantic labeling & \checkmark & & \checkmark & & & \checkmark & & \checkmark & & & & & & &\checkmark &\checkmark\\ \hline
Trajectory analysis & & & & & & & \checkmark & \checkmark & \checkmark & \checkmark & \checkmark & \checkmark & & & &\\ \hline
Image retrieval & & & & & & & \checkmark & & & & & & & & &\\ \hline
Data fusion & & & & & & & & & \checkmark & \checkmark & & & & \checkmark &\checkmark & \checkmark\\ \hline
Geospatial analysis & \checkmark & \checkmark & \checkmark & & \checkmark & \checkmark & & & & \checkmark & & & \checkmark & &&\\ \hline
Environmental monitoring & & \checkmark & & & \checkmark & & & & & & & & & &\checkmark&\checkmark\\ \hline
Disaster response & & \checkmark & & & & & & & & & & & & &\checkmark&\\ \hline
Remote sensing & & & \checkmark & & \checkmark & & & & & & & & \checkmark & \checkmark& &\checkmark \\ \hline
Compression & & & & \checkmark & & & & & & \checkmark & \checkmark & & & &&\checkmark\\ \hline
On-board satellite processing & & & & \checkmark & & & & & & & & & \checkmark & & &\checkmark\\ \hline
Movement prediction & & & & & & & & \checkmark & \checkmark & & & \checkmark & & &&\\ \hline
3D semantic segmentation & & & & & & & & & & & & & & \checkmark & &\\ \hline
Standardization & & & & & \checkmark & & & & & & & & & &&\\ \hline
High-resolution sensing & & & & & & & \checkmark & & & & & & & &\checkmark&\\ \hline
Vessel movement data analysis & & & & & & & & & & & \checkmark & & & &&\\ \hline
Urban mobility analysis & & & & & & & & \checkmark & & \checkmark & & & & \checkmark&&\\ \hline
Maritime logistics & & & & & & & & & & & & \checkmark & & &&\\ \hline
Cross-urban analysis & & & & & & & & & & & & & \checkmark & &&\\ \hline
Satellite data compression & & & & \checkmark & & & & & & & & & \checkmark & & &\checkmark\\ \hline
Deep feature optimization & & & & & & & \checkmark & & & & & & & \checkmark& &\checkmark\\ \hline
Stream analysis & & & & & & & & & \checkmark & & & & & &&\\ \hline
Mobility data analysis & & & & & & & & \checkmark & & \checkmark & & & & &&\\ \hline
GPS data processing & & & & & & & & & & & \checkmark & & & &&\\ \hline
Key component extraction & & & & & & & & & & & & \checkmark & & &&\checkmark\\ \hline
Handling diverse urban data & & & & & & & & & & & & & \checkmark & &\checkmark&\\ \hline
Water detection networks & & & & \checkmark & & & & & & & & & \checkmark & &&\\ \hline
Categorization for disaster assistance & & & & & & \checkmark & & & & & & & & &&\checkmark\\ \hline
\end{tabular}
\label{table1}
\end{table*}
\subsubsection{Computational semantic compression}
The study presented in  \cite{jia2013feature} investigates potential resolutions for constraints such as streamlining features, indexing, semantic labeling, and ultimately, approaches for formulating and executing searches. As stated in \cite{espinoza2013earth}, the primary objective of the study is to evaluate the existing condition of research on extracting information from EO data. The authors of \cite{augustin2019semantic} provide a new notion of a semantic EO data cube. It enriches traditional exploratory optimization data cubes by including an extra layer, referred to as a "nominal" or categorical interpretation, for each observation.  Thus, it becomes feasible to use semantic searches that directly relate to the substance of the data and are easily comprehensible to people.  By using machine-based reasoning, this approach improves automated analysis and information extraction in comparison to non-semantic cubes.  Fundamentally, semantic EO data cubes enable a wider range of users to utilize comprehensive EO data. This study assesses CORSA \cite{ivashkovych2022corsa} and \cite{beusen2022image}, a novel compression technique driven by artificial intelligence that produces valuable data representations. The authors contrast conventional water detection methods with innovative techniques that use the compressed data and representations generated by CORSA. The findings demonstrate that CORSA compression does not negatively impact accuracy and enables the direct establishment of water detection networks using its representations. The versatility of CORSA provides a viable option for the processing and compression of data onboard, tackling the growing difficulty of handling Earth Observation (EO) data. Nonetheless, compression is intrinsically application-specific, which poses challenges when striving to create diverse apps that share a similar dataset. In this context, the authors of \cite{tran2020semantic} suggest an innovative approach using semantic technology to establish a standardized language for Earth Observation data. This technique improves the integration of raster imagery data, such as land cover maps, with other geographical information, hence enhancing the efficiency of analysis and the use of this essential EO data.

The exponential growth of data produced by satellites and other remote sensing devices poses a new challenge: the rise of vast Earth observation data. Despite the great potential of this data for many applications, such as the provision of disaster assistance, its processing and interpretation present substantial difficulties. The study presented in \cite{li2021image} investigates existing methods for categorizing distinct images from this extensive collection, highlighting their practical usefulness. The authors examine potential methodologies for analyzing and managing large-scale EO data by using advancements in computer vision, machine learning, and other disciplinary domains. The study described in \cite{xia2017exploiting} investigates the many factors that might influence the effectiveness of deep features for image retrieval. By optimally adjusting these factors, the authors of \cite{xia2017exploiting} may get exceptional results on publicly available datasets. The present work provides valuable insights for future efforts in retrieving high-resolution remote sensing (HRRS) photographs by analyzing their content.
\subsubsection{Fusion of semantic trajectories}
Global Positioning System (GPS) data produces huge amounts of trajectory information. While scholars have largely focused on the preservation and analysis of these trajectories, there is a growing need to understand the underlying importance of the movement. The material includes traditional compression algorithms, the criteria for defining moving objects, approaches for measuring compression efficiency, and potential future applications. For real-time applications, conventional methods for producing substantial movement data are inadequately fast. The research presented in \cite{yan2013semantic} discusses a novel approach that combines mobility data with geography and application information to produce semantic trajectories. These improved trajectories not only reveal the destinations of objects but also provide a valuable understanding of the fundamental causes of their motion. 

The SeTraStream system, as described in \cite{yan2011setrastream}, tackles this problem by analyzing real-time semantic trajectories generated from live streaming movement data. This enables prompt processing for applications like location-based queries. Furthermore, the study shown in \cite{gao2019semantic} used an automated method to classify points in trajectories that display similar characteristics, leading to a more concise and meaningful depiction. This methodology is adequately effective in handling live data streams, making it suitable for real-time applications. An expanding array of advanced technologies described in \cite{richter2012semantic} and \cite{parent2013semantic} have become available for collecting and combining an unprecedented amount of data from numerous sources, including GPS trajectories related to the routes of mobile objects. Thus, the COMPRESS system described in \cite{han2017compress} is specifically developed to efficiently compress GPS data collected from urban road networks. The COMPRESS algorithm works by first partitioning the data into geographical routes and timestamps, then thereafter compressing each individual component independently. Similar to the method described in \cite{haiyan52vessel}, this technique achieves significant compression ratios while allowing location-based searches on the compressed data. Nevertheless, the authors of \cite{haiyan52vessel} compress the data by identifying and keeping just the most important components based on these criteria. This methodology effectively reduces the amount of data while maintaining crucial information for further study of ship movement patterns.
\subsubsection{Advancements in Semantic Comprehension}
To optimize the comprehension of movement data (trajectories) by incorporating semantic understanding, two crucial methods to consider are compression, which aims to decrease the volume of data, and semantic segmentation, which involves dividing trajectories into meaningful segments annotated with semantic labels. This paper focuses on emerging developments, including the integration of additional data dimensions and the facilitation of real-time processing. The preferred approach is contingent upon many elements, including urgent needs and the equilibrium between reducing the volume of data and comprehending its importance. Furthermore, the work outlined in \cite{liu2021semantics} presents a novel approach to integrating semantic information to enhance trajectory simplification. The approach does this by first distinguishing and delineating periods of movement from occurrences of halting within trajectories. Moreover, these sections are then simplified individually. The HighDAN AI network, as outlined in \cite{hong2023cross}, is built around this approach and has the capability to comprehend the significance of content conveyed by intricate imagery, such as satellite images, originating from various urban areas. The system exhibits remarkable competence in managing the wide range of ways in which these images depict identical geographical features in diverse urban settings. These capabilities allow HighDAN to attain exceptional performance and generalizability in the process of semantic compression from remote sensing data.

The research conducted in \cite{adam2023deep} directly investigates the use of semantic segmentation in three-dimensional meshes. Semantic segmentation is the process of categorizing different sections of a mesh into certain groups, such as urban structures or roadways. While deep learning has shown efficacy in assessing many types of remote sensing data, such as point clouds and satellite photos, there is a lack of assessments particular to 3D mesh segmentation. The present evaluation aims to fill the existing research gap regarding the use of deep learning techniques for the purpose of semantic segmentation of extensive 3D models inside urban environments. The authors explore the application of deep learning approaches to meshes, different ways of classification, and the assessment of experimental results \cite{adam2023deep}. The SAMRS framework, as described in \cite{wang2024samrs}, is a novel system that includes a very large collection of over 100,000 photographs. The photos undergo automated recognition and include exact EO data of superior quality. The SAMRS system utilizes existing object detection data and applies the Segment Anything Model (SAM) approach \cite{kirillov2023segment} to effectively provide semantic labels and segmentation. This dataset has many benefits, including its vast size, comprehensive item information, and adaptability for incorporating various semantic segmentation tasks.

Table \ref{table1} highlights the significant advancements in semantic image processing, computational semantic compression, and the fusion of semantic trajectories within the context of EO and related fields. From streamlining data searches and enhancing data integration to innovative compression methods like CORSA, these contributions emphasize the growing importance of semantic technologies in handling large-scale EO datasets. The development of semantic EO data cubes, real-time semantic trajectory processing systems, and advanced AI models such as HighDAN and SAMRS demonstrates how semantic understanding is transforming data analysis, enabling more efficient and accurate results. These innovations pave the way for improved applications across various domains, including autonomous vehicles, urban planning, and environmental monitoring. Nonetheless, challenges remain, particularly regarding non-standard data types and limited datasets, necessitating further research to optimize deep learning models and enhance their generalizability across diverse environments.
\section{Establishment of networks for semantic communication}
Due to its capacity to establish more effective communication systems, the use of semantic communication (SemCom), which in particular emphasizes the conveyance of the information's basic meaning (semantics), has grown in popularity. Despite this dedication, there is still a gap between the theoretical concepts and how they are really implemented. The authors of \cite{Utkovski2023} close the gap by reassessing the fundamental ideas of SemCom and how they affect system design. The investigation includes two primary components: a goal-oriented methodology that gives priority to metrics focused on data interpretation \cite{mostaani2022task}, and semantic operability, which considers both data sharing and its relevance. Within the field of engineering design, semantic networks play a crucial role by using knowledge graphs called nodes and links to visually represent ideas and their interconnectedness \cite{Han2021}. They are increasingly employed as storage places of information to support various engineering design tasks. 
\subsubsection{Semantic networks evolving toward SemCom networks}
An inherent attribute of artificially intelligent systems intended for various applications is their capacity to premeditate goal-oriented actions based on predetermined conditions of a problem space. The development of transformation, substance, and interpretation operations for semantic networks is aimed at organizing the behavioral planning of self-governing intelligent systems during the resolution of problematic issues \cite{melekhin2021fuzzy}. These procedures are supported by a formalization of the current situations in a problem space characterized by vast dimensions. Semantic networks aim to verify the understanding of a semantic representation and the ability to create or enhance a semantic language that corresponds to the viewpoint of the educational instructor. The engineering design field makes use of two main classifications of semantic networks:
\begin{itemize}
    \item WordNet \cite{miller1995wordnet} and ConceptNet \cite{liu2004conceptnet} are representative instances of general knowledge networks that serve as repositories for common-sense knowledge. These networks have the potential to be very beneficial for tasks such as idea generation and information retrieval. Nevertheless, they may be deficient in the requisite level of comprehensiveness for some technological fields.
    \item Engineering and technical networks, as documented in \cite{sarica2023design}, such as B-Link and TechNet, are specifically designed for engineering applications and include technical expertise and professional networks. Their capacity to improve efficiency in activities such as patent search and design review is considerable.
\end{itemize}
The restricted availability of infrastructure resources is a significant challenge in meeting the extensive demand with stringent performance standards for the goal of semantic networks. SemCom works by using information bases, which are collections of background and contextual data that are shared ahead of time between machines communicating to arrive at a shared understanding of the communication goal \cite{sana2022learning}. The SemCom networks enable a wide connection without requiring a common vocabulary or structure for data transfer, therefore improving communication effectiveness and reliability, resulting in higher Quality of Service (QoS) especially tailored to individual requirements.  

According to the SemCom networks outlined in \cite{shokrnezhad2024semantic}, a semantic-aware infrastructure enables the provision and orchestration of 6G services with rigorous Quality of Service (QoS) and Quality of Experience (QoE) standards. Service providers catalog their offers, including functional instances for activities like motion delivery and monitoring. The registration procedure is facilitated by pertinent Knowledge Bases (KBs), which may incur additional costs for computation, storage, and communication. Maintaining these knowledge bases necessitates computing resources for data updating and querying, storage capacity for information retention, and communication bandwidth for transmitting semantic data between nodes and service instances. To tackle these difficulties, services use lifelong learning initiatives to maintain the currency of knowledge repositories. Clients get these services by creating direct links that regulate the flow of semantic information to the service instances. In this scenario, the reasoning layer in terminal nodes, endowed with sophisticated linguistic skills, necessitates semantic representations from both user and control layers to enable various access instructions. This integration augments the cognitive abilities and operational independence of radio nodes, hence enhancing their decision-making processes for networking and control.

\subsubsection{Semantic communication perspective and characteristics}
In light of the limitations imposed by traditional methods, the SemCom systems of 6G need a more sophisticated approach to resource allocation, especially in dynamic environments. Conventional methods of communication sometimes overlook the importance of the fundamental relevancy of information. Their operation is based on the concept that each bit or symbol has equal significance and is considered as such. A key goal of these systems is to guarantee precise sequence retrieval at the receiving ends, with a focus on prioritizing the transmission of conformance. The predominant design techniques in this domain have been established based on ideas derived from digital communication. Theoretical limits of information determine the highest possible capacity of a certain system. The goal of channel coding is to devise techniques that can achieve these restrictions with a very low probability of errors. Conversely, source coding, often known as data compression, involves encoding data in a manner that reduces the necessary data size to maximize the length of the encoded sequence from the source. Nevertheless, the latest manifestation of communication systems is being executed in manners that question the traditional design paradigm, particularly with regard to semantic elements. 

In \cite{huang2023large}, deep reinforcement learning is used to tackle the challenges encountered in multi-cell systems. By translating data rates into semantic information rates and utilizing a unique metric called Semantic Throughput (STP), the technique improves resource allocation and achieves improved STP performance compared to standard methodologies. This not only improves the effectiveness of communication, but also guarantees the dependable transmission of meaning, therefore laying the foundation for advancements in 6G. As Shannon's Law nears its ultimate limits, researchers are exploring the theoretical constraints of compressing substantial information (semantic information). The Bayesian network used to represent the semantic source evaluates both lossless and lossy compression techniques \cite{Xia2024}. The authors establish theoretical limits for attaining perfect reconstruction and propose an efficient encoding model that depends on the network's structure. Moreover, they provide the most efficient compression methods for scenarios where information is available to both the sender and receiver. They illustrate that the optimal method entails incorporating distinct enhancements throughout the network. This work provides a basis for understanding the maximum degree of compression that may be successfully achieved for semantic data.

The semantic attribute assesses the congruence between the recipient's deduced understanding and the presenter's intended message, by considering the content, specifications, and semantics, with the aim of improving the intelligence of the communication system. Coherence encompasses both the precision and lucidity with which symbols may convey their intended meanings, as well as the effective conveyance of information or a concept from its origin to its destination, without delving into particular aspects. The evaluation of the compression level attained in semantic source coding in relation to the original information is of utmost significance \cite{tang2023informationtheoretic}. Through the use of this semantic source, the authors have derived the theoretical limits for both lossless and lossy compression, along with the minimum and maximum constraints on the rate-distortion function.  Based on the defined research objectives and existing knowledge, it can be shown that semantic source coding \cite{Lu2022SemanticsEmpoweredCA} has the potential to achieve more significant reductions in communication overhead and redundancy cost when compared to separate source-channel coding (SSCC). 
\subsubsection{Semantic joint source-channel coding}
\begin{figure*}[htbp]
\centerline{\includegraphics[width=170mm]{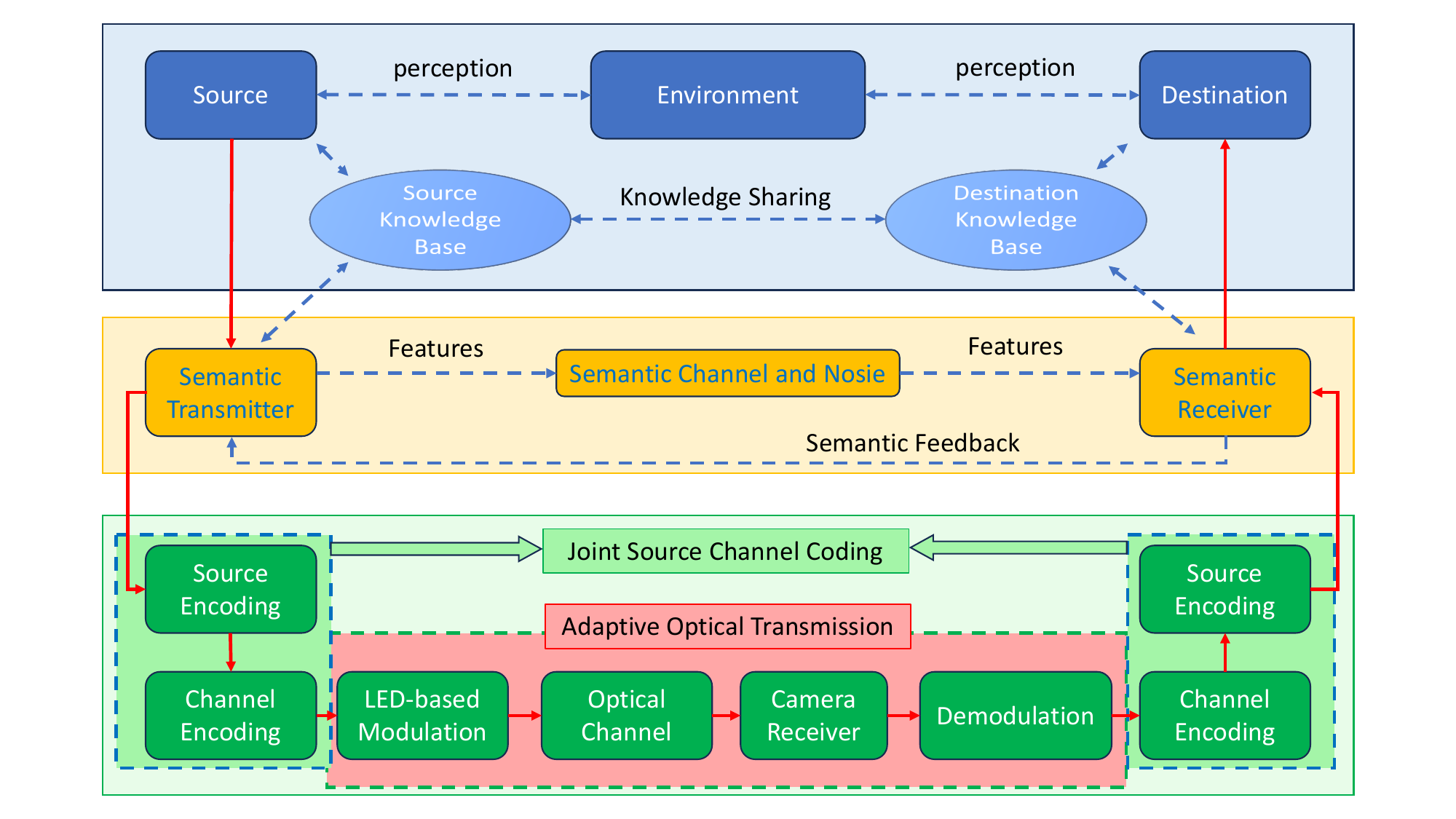}}
\caption{The block diagram for the proposed semantic inference-based earth observation adapting the techniques in \cite{chen2024semantic}.}
\label{Systemblock}
\end{figure*}
Traditional communication networks treat all data as equal and provide priority to sending it without any mistakes. This statement neglects to recognize the varying importance of the information itself. Source coding is a data compression technique used to prioritize the delivery of just the essential components. This suggests that conventional methods may be improved by giving priority to the relevancy of information. The objective of SemCom is to understand the fundamental meaning of the message. Despite its growth, the discipline has challenges such as developing models for complex data and using current communication systems.  Notwithstanding the obstacles, the JSCC method has the capacity to enhance semantic communication in future systems like 6G.

The partnership between JSCC and deep learning, as presented in \cite{farsad2018deep}, showcases that the deep learning encoder and decoder outperform the traditional approach in terms of word error rate, particularly when there are limitations on the computational resources designated for syllable encoding. A fundamental limitation of this approach is the use of a predetermined bit size to encode phrases with varying lengths. Analysis of the deep semantic communication (deep-SC) scheme's complexity, as shown in \cite{Xie2021}, reveals its superior performance compared to existing SSCC schemes. In contrast, the semantic communication technique based on JSCC exhibits reduced processing time compared to the deep-SC system. The successful incorporation \cite{Xie2021} and analysis \cite{Lu2022SemanticsEmpoweredCA} of deep learning and JSCC techniques provide a valuable motivation for the innovative advancement of semantic communication. The fundamental objective of this work is to analyze the process of semantic source coding in connection to primary information. The objective of this framework is to surpass the constraints of Shannon's conventional information theory, therefore enabling a more all-encompassing methodology. 

The primary goal of deep learning-based Joint Source-Channel Coding (DJSCC) \cite{tung2022deep} is to reliably attain high levels of performance, even in scenarios characterized by resource constraints and low Signal-to-Noise Ratios (SNRs). The reconstruction scheme involves the use of DJSCC, as outlined in the research by \cite{kurka2021bandwidthagile}, which may be seen as an adapted version of an auto-encoder. Supplementary noise ratios are included in the intermediate part of the encoder, resulting in the generation of a modified version of the original auto-encoder. In contrast to traditional wireless transmission techniques, the DJSCC encoder \cite{Kurka2022} employs semantic signals, which are analog waveform sequences, instead of digital transmission. Consequently, it may operate with high efficiency even in challenging channel conditions while still achieving enough recovery. Nevertheless, the enhanced efficiency attained by modern DJSCC techniques is only evident via simulations in scholarly research, which function as a type of information dissemination via supplementary transmission for semantic networks \cite{Shi2023}. Typically, these models make the assumption of flawless synchronization, precoding, antenna, and power amplifier conditions. Therefore, the study \cite{Liu2022} focuses on a DJSCC platform that employs software-defined radio technology. The evaluation of the system's capability relies on two crucial factors, namely non-linear distortion and synchronization inconsistencies.  Moreover, the authors of \cite{Yoo2023} propose an innovative method for semantic communication, drawing inspiration from the impressive adaptability shown by Vision Transformers (ViTs) in successfully addressing various issues associated with image disruptive behavior. The potential advantages outlined in \cite{Zhong2024} of incorporating Unequal Error Protection (UEP) into a semantic encoder/decoder, using the current JSCC system, are worth investigating. The objective of this approach is to preserve the relative importance of data focused on semantic tasks. Furthermore, the implementation of a JSCC system may be simplified by using Quasi-cyclic Low-Density Parity-Check (QC-LDPC) codes on an adaptable device. A prototype for semantic communications is developed by constructing a fixed-point system that is then used to transmit and receive semantic information across a channel that contains residual noise. 

In a recent publication, Zhang presented a new training approach for semantic communication systems. This approach builds upon the research accomplished by \cite{zhang2024analog} and demonstrates its superiority compared to current digital and analog JSCC methods. A key challenge is in including modulation, a crucial step for effective transmission, into the training process. In order to address this problem, the authors use a theoretical framework that views modulation as a technique for data compression and employs noise to simulate its effects. This methodology enables the teaching of real-life situations and much exceeds current approaches in terms of efficacy. Two innovative approaches are proposed in the research of \cite{yang2024digital} to facilitate the seamless integration of digital communication and DJSCC. The first modulation, uniform modulation, functions as a flexible spring by modifying the spacing between data points (constellations) during transmission to align more effectively with the unique properties of the picture data being sent. The second technique, known as non-uniform modulation, adopts a more customized approach. The system examines the picture data and generates constellations which are particularly tailored to the distinct distribution of that particular image, therefore guaranteeing maximum transmission efficiency.  Both methodologies ingeniously tackle the transformation of continuous analog data into discrete digital signals by meticulous adjustment of the system. The findings are remarkable, as this approach far surpasses current ways, thus facilitating enhanced picture transfer in digital communication. The work described in \cite{huang2024d2jscc} introduces a novel approach called D$^2$-JSCC to address the discrepancy in digital sources and thereby unlock the full potential of DJSCC. D$^2$-JSCC operates as a tag team, where one component of digital source coding analyzes the image and identifies its semantic properties, taking into account the distribution of this statistical information. 

The integration of semantic technologies in the fast-growing area of communication systems offers a novel method for improving data transmission efficiency. Our proposed technique presents a semantic transmitter that collects and encodes significant information directly from the source, enhancing the data preparation for transmission. This strategy utilizes the distinctive characteristics of light-emitting diodes (LEDs) in optical communication to enhance information transmission \cite{chen2024semantic}, especially in situations where conventional approaches may be hindered by noise and bandwidth constraints. The semantic transmitter functions by examining the content and context of the data before modulation. This facilitates the extraction of pertinent elements that enhance the display of information, hence improving decision-making for the recipient. The modulation method then converts these semantic representations into a format appropriate for LED-based optical transmission, assuring the preservation of the information's integrity and relevance throughout the communication process.

As illustrated in Fig.\ref{Systemblock}, a semantic transmitter extracts and encodes meaningful information from the source, which is then modulated using light-emitting diode (LED)-based optical transmission \cite{chen2024semantic}. The optical channel carries this data to a camera receiver, where it is demodulated and decoded through JSCC to recover the original semantic information. A key feature of the system is its use of feedback loops and knowledge sharing, which allow the transmitter to adapt its encoding and transmission strategies in response to channel conditions, improving robustness and efficiency. The system employs adaptive optical transmission, adjusting parameters like modulation and power levels to enhance performance under varying environmental conditions. The knowledge base in this semantic communication system plays a central role in enhancing efficiency and accuracy by focusing on transmitting meaningful information rather than raw data. The transmitter and receiver can interpret semantic representations by sharing a common knowledge base, ensuring efficient communication with reduced data transmission. The system incorporates real-time feedback, allowing the transmitter to adapt its encoding and transmission strategies based on current channel conditions, such as noise or interference. 

Incorporating an LLM enhances the system's ability to understand and generate contextually relevant content, improving the semantic accuracy of the transmitted information. Real-time feedback mechanisms allow the transmitter to adapt its encoding and transmission strategies based on current channel conditions, such as noise or interference. This adaptive approach optimizes LED-based optical transmission and employs joint source-channel coding, focusing on high-level semantic features instead of relying solely on traditional error correction methods. The knowledge base, augmented by the LLM, enables the system to intelligently manage noise and reconstruct lost data, resulting in a more bandwidth-efficient and power-optimized communication process, especially in dynamic environments. By leveraging the semantic understanding provided by the LLM, the system can deliver robust communication that aligns closely with user intent and context.

\section{Enhancing Domain-Specific Semantic Effectiveness with Large Language Models}
The development of LLMs has sparked significant interest in their ability to adapt to various domains. However, challenges remain, particularly in the areas of cognitive interference and memory retention when models are customized for specific tasks. Furthermore, the quest for a versatile LLM that can operate across multiple domains often leads to performance declines due to communication issues. To enhance LLM capabilities, integrating data fusion techniques becomes essential. Data fusion enables the synthesis of information from multiple sources, leading to richer context and improved decision-making. This integration can significantly enhance the model's understanding and responsiveness in complex environments, such as networking and Earth observation. This discussion explores recent strategies and frameworks aimed at optimizing LLM performance, highlighting innovative approaches to enhance their adaptability, efficiency, and semantic processing through effective data fusion.

\subsubsection{Domain-adapted large language model}
LLMs have difficulties in gaining various talents without encountering cognitive interference or memory deterioration when customized for particular domains.  Furthermore, the endeavor to construct a versatile model capable of handling many domains simultaneously often results in a decline in overall performance due to the presence of miscommunication across the various domains. REGA presented in \cite{wang2024role} has three strategies to tackle this issue: firstly, it uses illustrative examples to refresh the LLM's memory of general information; secondly, it provides particular instructions for each new skill via customized prompts; and thirdly, it seamlessly incorporates a limited quantity of task-specific material with the existing general knowledge. However, creating customized deep-learning models for individual networking tasks is a laborious process that also faces difficulties when encountering novel scenarios. The authors of \cite{wu2024large} propose using extensive LLMs that have been pre-trained on massive datasets. The NetLLM method, which follows a "one model for all" strategy, seeks to address diverse networking issues by enhancing performance and adapting to new circumstances. Another factor to be taken into account is that a scarcity of labeled data impedes the training of text categorization algorithms.  Domain Adaptation Large Language Model interpolator (DALLMi) presented in \cite{bețianu2024dallmi} employs a novel approach by using a small quantity of labeled data with a substantial volume of unlabeled data. It connects the two via word embeddings and a method known as interpolation. DALLMi additionally adjusts the training process to compensate for the uneven distribution of data. This strategy demonstrates superior performance compared to conventional approaches, particularly in situations when there is a limited availability of labeled data.  
\subsubsection{Design of semantic effective plane}
Although there has been notable progress in the development of LLMs, the field of semantic processing, which focuses on comprehending the meaning in language, seems to be experiencing a lack of significant innovation.  By processing natural language input, it understands context, identifies key information, and generates coherent, contextually appropriate actions or solutions. For example, when asked to summarize a document, it doesn't just extract random sentences—it analyzes the entire content, synthesizes the core ideas, and provides a concise summary. Similarly, when instructed to provide a recommendation, it evaluates the relevant data, processes the underlying patterns, and offers an informed suggestion. The key strength of a language model lies in its ability to turn complex text into clear, actionable outcomes, bridging the gap between human language and practical tasks. 

OpenAGI \cite{ge2024openagi} is an open-source Artificial General Intelligence (AGI) research platform designed to solve complex, multi-step tasks by synthesizing diverse models and utilizing datasets from sources like Hugging Face, GitHub, and LangChain. It introduces the Reinforcement Learning from Task Feedback (RLTF)  approach, where a Large Language Model acts as a controller to select and execute external expert models, using task feedback to refine its planning strategy. By combining LLMs with domain-adapted models, OpenAGI aims to enhance task-solving capabilities, even with smaller-scale models, advancing the potential for AGI. This is achieved by analyzing existing studies on these tasks and their practical applications, as illustrated in Fig.\ref{LLM} which  depicts the semantic effective plane of Fig.\ref{SystemModel}. Our proposed semantic effective plane can be achieved by the following outlines:
\begin{figure*}[htbp]
\centerline{\includegraphics[width=\textwidth]{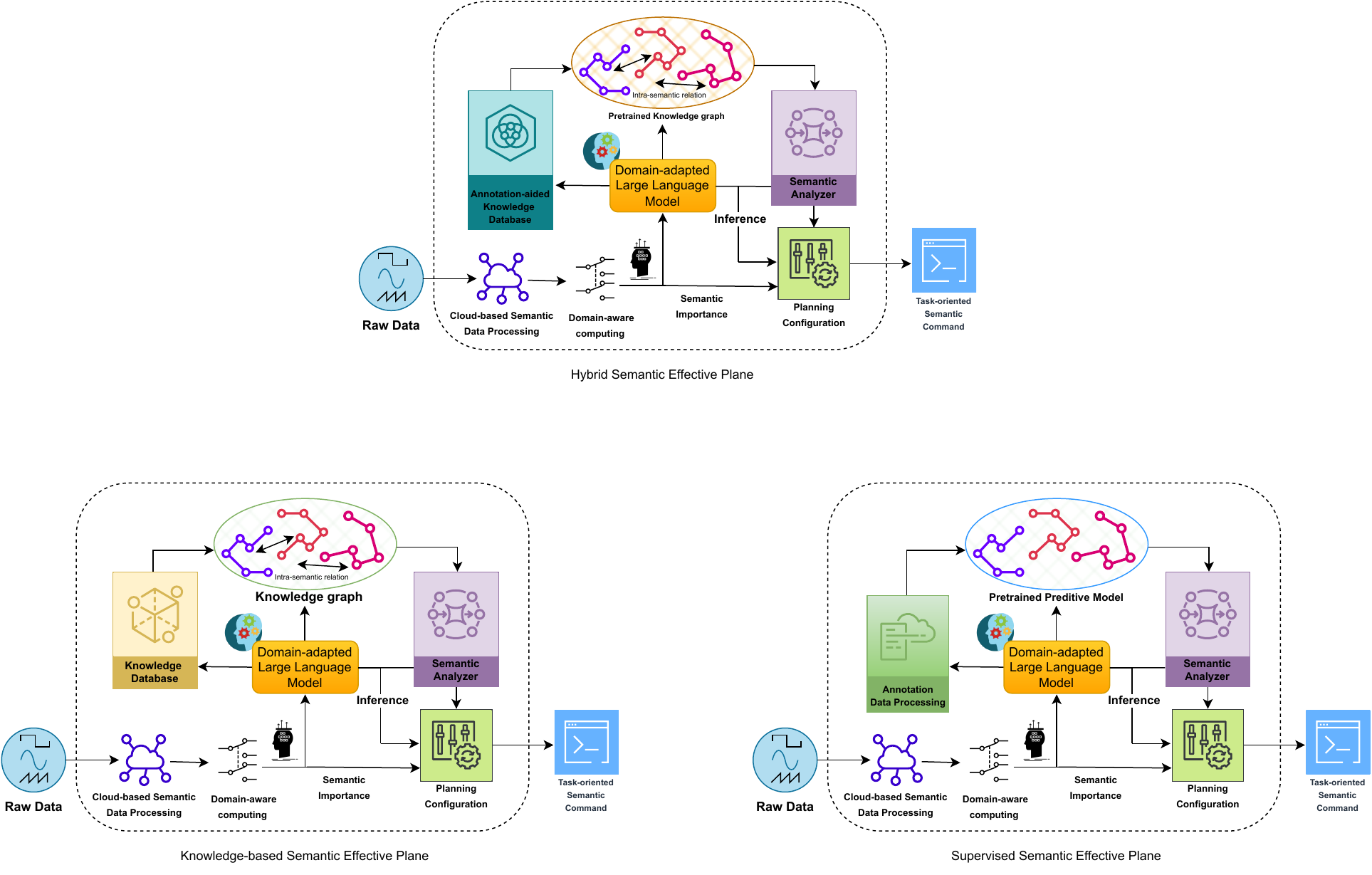}}
\caption{The proposed semantic effectiveness plane applying semantic processing with RAG LLM technique}
\label{LLM}
\end{figure*}
\begin{itemize}
    \item  By leveraging knowledge graphs, OpenAGI enriches its understanding of complex tasks through a structured representation of entities and their relationships, allowing for more accurate and context-aware decision-making. Integrating Retrieval-Augmented Generation (RAG) \cite{gao2023retrieval} enhances this approach by enabling OpenAGI to pull in relevant external information from large-scale databases dynamically, supplementing its internal knowledge base with the latest domain-specific insights. This combination of knowledge graph structuring and RAG-powered information retrieval facilitates a deeper comprehension of the domain-specific knowledge and connections relevant to multi-step tasks, allowing OpenAGI to adapt to evolving contexts and deliver precise, informed responses across varied applications
    \item In conjunction with knowledge graphs, OpenAGI utilizes pretrained predictive models to forecast potential outcomes and suggest optimal actions based on historical data and learned patterns. These models, trained on extensive datasets, provide valuable predictions that guide the LLM in selecting and synthesizing the most appropriate expert models for each task.
    \item The semantic analyzer within OpenAGI extracts semantic importance from natural language queries and task descriptions. This capability enhances the platform's ability to discern and prioritize critical elements within the text, ensuring that the most relevant information drives the task-solving process. By combining these components, OpenAGI not only synthesizes diverse models but also benefits from a deeper contextual understanding and predictive accuracy, advancing its potential for achieving AGI.
    \item The hybrid mode of knowledge graphs and pretrained predictive models empowers OpenAGI to tackle Earth Observation challenges from a comprehensive perspective, integrating domain knowledge with predictive insights. This integration enables OpenAGI to refine its task-solving strategies through Reinforcement Learning with Task Feedback (RLTF), leading to enhanced accuracy and efficiency in addressing complex, real-world scenarios. 
\end{itemize}
OpenAGI represents a significant advancement in the integration of artificial intelligence and EO technologies. By utilizing a hybrid approach that combines knowledge graphs with pretrained predictive models, OpenAGI can tackle complex environmental challenges with greater precision and relevance. For instance, OpenAGI can generate task-oriented semantic commands such as "identify and classify land cover types in the provided satellite imagery" or "detect and monitor changes in water bodies over time." These actionable commands guide the execution of appropriate models or tools, ensuring that the system effectively leverages both the knowledge from the knowledge graph and the predictive capabilities of pretrained models. By applying this approach, OpenAGI can enhance tasks such as disaster response planning, resource management, and environmental monitoring, ultimately contributing to more informed decision-making in Earth Observation applications. 
\subsubsection{Automating earth observation with LLMs integration}
Domain-adapted LLM is promising to transform Earth observation by automating and improving analysis processes, making results more accessible to decision-makers. The framework presented in \cite{wagner2024generative} leverages LLMs to interpret and orchestrate tasks using tools such as the Open Data Cube and OEA Algorithm Hub, effectively streamlining EO analysis and showcasing the potential of LLMs in enhancing geospatial data processing and decision support. 
GeoLLM-Engine harnesses the power of natural language to execute complex EO tasks by integrating geospatial API tools, dynamic maps, and multimodal knowledge bases \cite{singh2024geollm}. Unlike existing systems that rely on simplified tasks and template-based prompts, GeoLLM-Engine evaluates agent performance on realistic, high-level commands through a massively parallel engine using 100 GPT-4-Turbo nodes, handling over half a million multi-tool tasks and 1.1 million satellite images. This approach advances beyond traditional single-task methods to assess state-of-the-art agents and prompting techniques with long-horizon prompts.
\section{On-board deep learning and hardware design for resource-efficient EO systems}
This section introduces the deep learning models for on-board processing are revisited based on complexity and efficiency, focusing on metrics like model size, parameters, and training/inference time. Furthermore, recent advancements in on-board processing hardware, including SRAM-based FPGAs, radiation-hardened processors, and reconfigurable MPSoCs, have enabled more resilient and adaptable computing for space missions. These systems enhance task performance and mitigate radiation effects, ensuring reliable operation even in harsh space environments. Collectively, these technologies support efficient data handling and communication, aligning with modern satellite objectives and expanding possibilities for advanced space exploration applications.

\subsection{Revisit on the effectiveness of on-board EO deep learning}
A detailed comparison of several deep learning models is presented in \cite{le2024board} focusing on metrics like input size, total parameters, estimated model size, FLOPs, and the time taken for training and inference. Simpler architectures, such as CNN and ResNet-50, exhibit low computational complexity. For instance, CNN has the smallest total parameters (66,330) and FLOPs (0.93 MFLOPs), making it highly efficient for training (233 seconds) and inference (7 seconds). ResNet-50, while more complex, still maintains moderate efficiency, with 0.2 million parameters, a model size of 10 MB, and 117.88 MFLOPs, resulting in faster inference times (6.7 s) compared to more complex models. Transformer-based models, such as ViT, SwinTransformer, and EfficientNet variants, are far more computationally intensive. ViT-large (Google) leads the complexity spectrum with 303 million parameters, 59.7 GFLOPs, and significant memory requirements (1.64 GB), translating into much longer training (9,271 s) and inference (60 s) times. SwinTransformer and EfficientNet-V2 models offer a compromise between complexity and efficiency. SwinTransformer, for instance, has 86.7 million parameters, 21.47 GFLOPs, and takes around 538 seconds for inference, making it efficient for high-performance tasks despite its complexity. The authors presented in \cite{le2024board} highlight that transformer models like ViT-huge (Google) and ViT-large are the most resource-intensive, while smaller models like SmallViT and EfficientNet-V2 maintain a good balance between model size, computational complexity, and time efficiency. 

Ultimately, this comparison shows a trade-off between performance and computational demand, with traditional CNN models being more efficient for simpler tasks, and transformer models excelling at complex, large-scale tasks but requiring significantly more computational resources.

\subsection{On-board processing hardware for EO semantic communication}
Telecommunication satellites are facing the need to meet the growing demand for faster data rates, smaller antenna sizes, and reduced power consumption in mobile devices. This necessitates the development of a streamlined bridge that is specifically designed for the processing technologies used in the application \cite{Hofmann2012}. On satellites designed for specific purposes or with a limited lifespan, there is some utilization of on-board processing.

\subsubsection{SRAM-aided programmable on-board processing}
The Cibola Flight Experiment (CFE) presented in \cite{quinn2015cibola} assessed the feasibility of using SRAM-based Field-Programmable Gate Arrays (FPGAs) for in-flight data processing in spacecraft. Introduced in 2007, it transported three adaptable computer modules, each equipped with Xilinx Virtex XQVR1000 FPGAs for the purpose of data processing.  The objective of the experiment was to analyze the impact of radiation on these Field-Programmable Gate Arrays (FPGAs) and assess methods to minimize or counteract these effects. The success of the CFE indicates that SRAM-based FPGAs have promise for future uses in spaceflight. In addition, SRAM-based Field-Programmable Gate Arrays (FPGAs) have shown their worth in performing crucial functions in the field of space exploration \cite{quinn2015cibola}. Xilinx XQR4062XL FPGAs were responsible for managing the descent and landing of the Mars rovers Spirit and Opportunity.  These missions aimed to overcome the susceptibility of SRAM-based FPGAs to space radiation by using fault mitigation mechanisms. This included implementing Triple Modular Redundancy throughout the design phase and including redundant Field-Programmable Gate Arrays as components. The FPGAs have the capability to be fully reprogrammed in order to effectively eliminate any faults. Although redundancy provides resilience, it also incurs additional weight, power use, and total expenditure. 

The Fraunhofer aboard-Board Processor (FOBP) presented in \cite{rittner2014broadband} takes a distinct approach by using radiation-hardened Virtex-5QV FPGAs to provide adaptable communication processing aboard satellites. The main benefit of FOBP is its capacity to be reconfigured while in orbit. By reprogramming the FPGAs from ground stations, the processor can adapt to new communication protocols and changes in the environment, going beyond the limits of traditional systems that can not be changed. This opens up the possibility of using non-radiation-protected FPGAs such as Kintex UltraScale or Zynq-UltraScale+ in Low Earth Orbit (LEO) space applications \cite{hofmann2017reconfigurable}. A method improves onboard reconfiguration of SRAM-based FPGAs in LEO satellite networks \cite{Qiao2023}. This approach automates the process, breaking down data for transfer, verifying temporary storage, and ensuring successful flash memory programming. Built-in error checks throughout this closed-loop system enhance reliability, allowing for flexible and dependable maintenance of spacecraft FPGAs.

\subsubsection{Multiprocessor systems-on-chips on-board processing}
Multiprocessor Systems-on-Chips (MPSoCs), which integrate CPUs, FPGAs, and other components into a single chip, provide distinct adaptability for space applications. The reason for this is that they integrate the flexibility of software with the programmability of FPGAs. This enables the construction of radiation-resistant systems by using a variety of these structures.  An example of this methodology is the CHREC Space Computer (CSP). The CSP presented in \cite{rudolph2014csp} employs a Xilinx Zynq platform and a blend of software and hardware hardening methods to attain exceptional dependability, as confirmed by space missions.  Like the CSP, the APEX-SoC system similarly uses a single Zynq device for gathering and analyzing data \cite{iturbe2015integrated}. 

The potential of reconfigurable MPSoCs is applicable even to space agencies, such as the European Space Agency's OPS-SAT program \cite{garcia2021remote}. This program places a high priority on using commercially available technologies (COTS) that are already in existence to develop a satellite that is adaptable, allowing for the interchangeability of applications even after it has been launched. The OPS-SAT payload utilizes the ALTERA Cyclone V processor as its core, whereas the NINANO platform \cite{bezerra2017evaluation} employs a Xilinx Zynq-7030 for on-board processing in scientific missions such as EYE-SAT which involves imaging of zodiacal light \cite{levasseur2014eye}. The authors of \cite{Perez2020} present a specialized on-board CPU architecture that is adaptable and intended for crucial space applications. The primary characteristic is a multi-accelerator hardware system that has the ability to be reprogrammed in real-time while in operation. This enables adaptable optimization of performance, dependability, and energy use according to the requirements of the task.  The architecture seamlessly integrates with a run-time reconfigurable operating system, ensuring adherence to space mission requirements and enabling both reconfiguration and failure mitigation. For example, the authors employed the on-board processor to create a vision-based navigation system. The system has the ability to autonomously adapt to different stages of a mission or unexpected situations, highlighting the practical benefits of reconfigurable MPSoCs for real-world space missions. In addition, the research presented in \cite{pagonis2023increasing} indicates that the use of Digital Signal Processors (DSPs) greatly enhances the ability to handle errors while performing crucial calculations, in contrast to the use of Look-Up Tables (LUTs). This results in a significant decrease in downtime, with reductions of 95\% and 65\% for certain activities, thus reinforcing the argument for using reconfigurable MPSoCs in space applications. 

Cost-effective solutions are frequently necessary for new space missions. The authors of \cite{panousopoulos2024hw} aim to enhance the efficacy of star trackers in satellites, which is crucial for preserving orientation.  They accomplish this by employing reconfigurable MPSoC to accelerate image processing algorithms, which is a more cost-effective method than conventional methods.  Their accomplishments were remarkable: they were able to accelerate computations by 95\% and process 4-megapixel images at a rate of over 24 frames per second through the use of ingenious design techniques.  This optimization ensures that the system operates in real-time and remains accurate, rendering it a viable option for space programs that prioritize cost.
\begin{figure*}[htbp]
\centerline{\includegraphics[width=\textwidth]{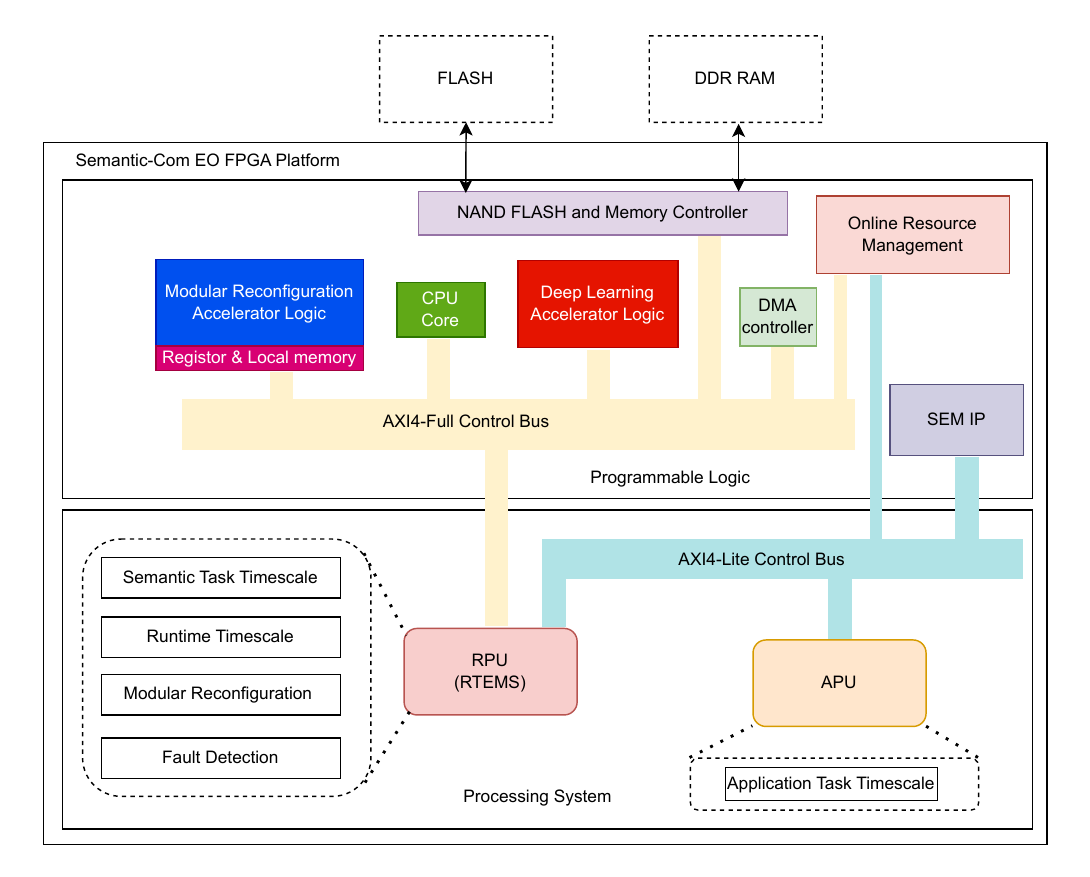}}
\caption{The proposed EO on-board processing for semantic communication inspired by \cite{Perez2020}.}
\label{EOOP}
\end{figure*}
\subsubsection{Runtime and modular reconfiguration on-board processing}
The avionics and space industries are experiencing an increase in the popularity of on-board processors. This technology, which was first implemented on the FedSat mission \cite{fraser2003fedsat}, is currently employed to resolve errors and modify functions in mid-flight, as seen in Cibola. This dual-purpose reconfiguration is advantageous for electronics that operate in challenging environments. This research establishes a spacecraft system that is adaptable by utilizing the reconfigurable ARTICo$^3$ architecture \cite{rodriguez2018fpga}. ARTICo$^3$ enables the hardware to be reprogrammed during flight to ensure fault tolerance and adaptable performance. The system seamlessly integrates with a space-ready operating system (RTEMS) and includes error-handling capabilities. Additionally, it facilitates both mission-specific duties and general tasks (e.g., defect management and reconfiguration). For instance, the researchers of \cite{rodriguez2018fpga} demonstrated the system's adaptability by incorporating vision processing that adjusts to various mission phases. This method illustrates a promising avenue for the development of spaceborne systems that are both flexible and dependable. It is worth noting that ESA's study presented in \cite{leon2021development} expands the limits of on-board partial reconfiguration using the BRAVE platform developed by NanoExplore \cite{oliveira2019dynamic}.  BRAVE FPGAs are purpose-built for space applications, with a radiation-resistant material that ensures their reliability in challenging situations.  An important advancement is the use of a SpaceWire interface for reconfiguration, which is a specialized space communication standard that may provide benefits compared to conventional JTAG approaches as investigated in an independent study \cite{bravhar2018brave}. This exemplifies the continuous progress in reconfigurable on-board processing for versatile and reliable spacecraft.

As illustrated in Fig.\ref{EOOP}, we propose implementing an on-board processing for semantic communication EO  based on the research the study presented in \cite{Perez2020} using a Zynq UltraScale+ MPSoC. The system integrates versatile Application Processing Unit (APU) to handle a wide range of activities, dedicated Real-Time Processing Unit (RPU) for time-sensitive operations, and customizable reconfigurable Programmable Logic (PL).  Real-Time Executive for Multiprocessor Systems (RTEMS), a real-time operating system, effectively controls all tasks and guarantees prompt execution including semantic task schedule. The RPU and APU operate the essential application of the satellite, such as navigation, in addition to performing support activities including managing errors, controlling reconfigurations, and communicating with ground base station. This platform focuses on enhancing the fault tolerance of spacecraft computers via the use of a specialized Cortex-R5 processor-based RPU. The RPU utilizes lock-step execution, which involves completing computations twice in order to identify mistakes. The attainment of reconfigurability is made possible by the use of the Zynq UltraScale+ MPSoC hardware component, which enables adjustments to be made during missions. The Xilinx Soft Error Mitigation (SEM) IP proactively detects and rectifies problems induced by radiation. The RPU functions as the central hub, overseeing both reconfiguration and error correction. The fault-tolerant design incorporates redundant processing, modular hardware, and on-chip correction to enable dependable and adaptive spaceborne computing. Efficient communication is achieved via the use of separate AXI4 busses. The use of distinct AXI4 busses enables effective communication within this resilient design. The integration of redundant processing, adaptable hardware featuring deep learning accelerator capabilities, and on-chip error correction enables the development of resilient and flexible computing platforms for spacecraft. This platform is essential for ensuring the reliable execution of deep learning or other resource-intensive tasks in challenging space environments.

\subsection{Numerical Results for on-board EO deep learning}
Digital Task-oriented Joint Source-Channel Coding (DT-JSCC) \cite{xie2023robust} enables the seamless integration of digital communication and DJSCC. The initial modulation, known as uniform modulation, operates like an adaptable spring, adjusting the spacing between data points (constellations) during transmission to better align with the specific characteristics of the transmitted image data. This is the most suitable option for semantic communication in the aforementioned circumstance.
\subsubsection{Physical layer channel model}
The signals that were transmitted and received are denoted by $X$ and $Y$, respectively. The general form of the channel equation is $Y = H \cdot X + N$, where \( H \) is the channel gain matrix and \( N \) is the noise term. Below are specific formulae for different sorts of channels. $Y = X + N_{\text{AWGN}}$, where $N_{\text{AWGN}} \sim \mathcal{N}(0, \sigma^2)$, because $H = 1$ due to the absence of fading and has a Gaussian noise variance of $\sigma^2$. However, Rician distribution characterizes the channel gain $H$. The expression $Y = H_{\text{Rician}} \cdot X + N_{\text{AWGN}}$ is Rician-distributed, and the Rician $K_r$-factor defines $H_{\text{Rician}}$.

\subsubsection{Simulation results using EuroSAT dataset}
\begin{figure}[!ht]
    \centering
    \includegraphics[scale=0.6]{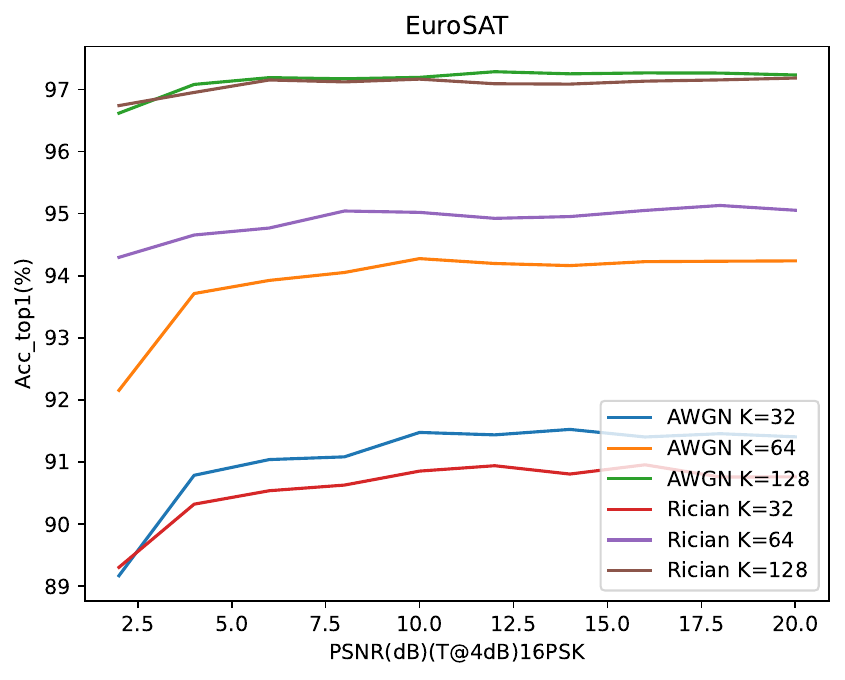}
    \includegraphics[scale=0.6]{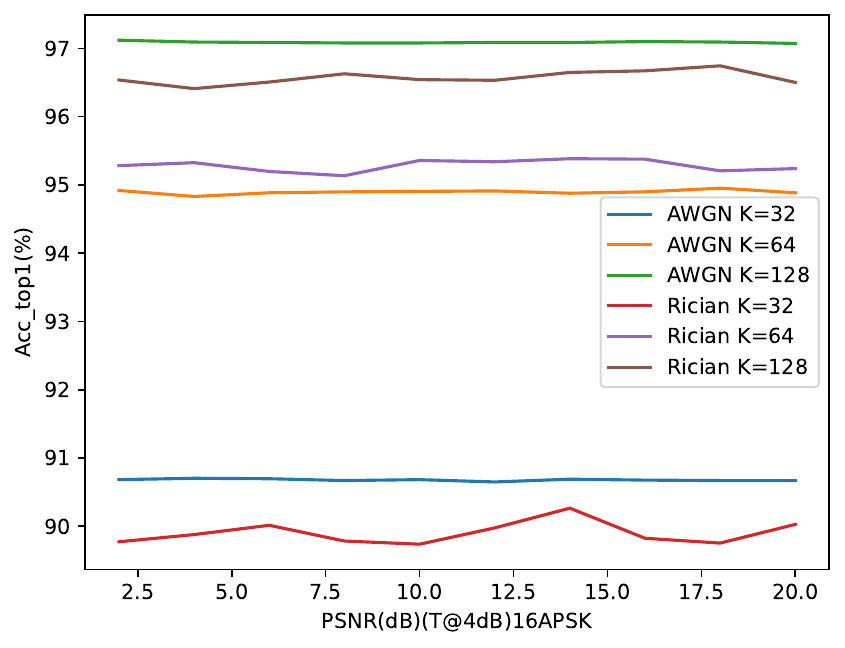}
    \caption{Top1 accuracy using DT-JSCC based on 16PSK/16APSK AWGN and Rician channel.}
    %\caption{Top1 accuracy using DT-JSCC based on 16PSK AWGN and Rician channels.}
    \label{fig:top1}
\end{figure}

Top-1 accuracy utilizing DT-JSCC for EuroSAT data under various channel circumstances is shown in Fig. \ref{fig:top1}. The performance of the 16PSK/16APSK modulation technique at various PSNR (Peak Signal-to-Noise Ratio) levels is shown in Fig. \ref{fig:top1}. With varying K-factor values (32, 64, 128), we demonstrate accuracy for the Rician and AWGN (Additive White Gaussian Noise) channels; in general, a larger K-factor indicates better channel quality. Additionally, for these K values, we compare accuracy in the Rician channel. The figures show that accuracy increases with rising PSNR in all conditions, that AWGN channels outperform Rician in general, and that greater K-factors result in more steady performance.

\begin{figure}[!ht]
    \centering
    \includegraphics[scale=0.38]{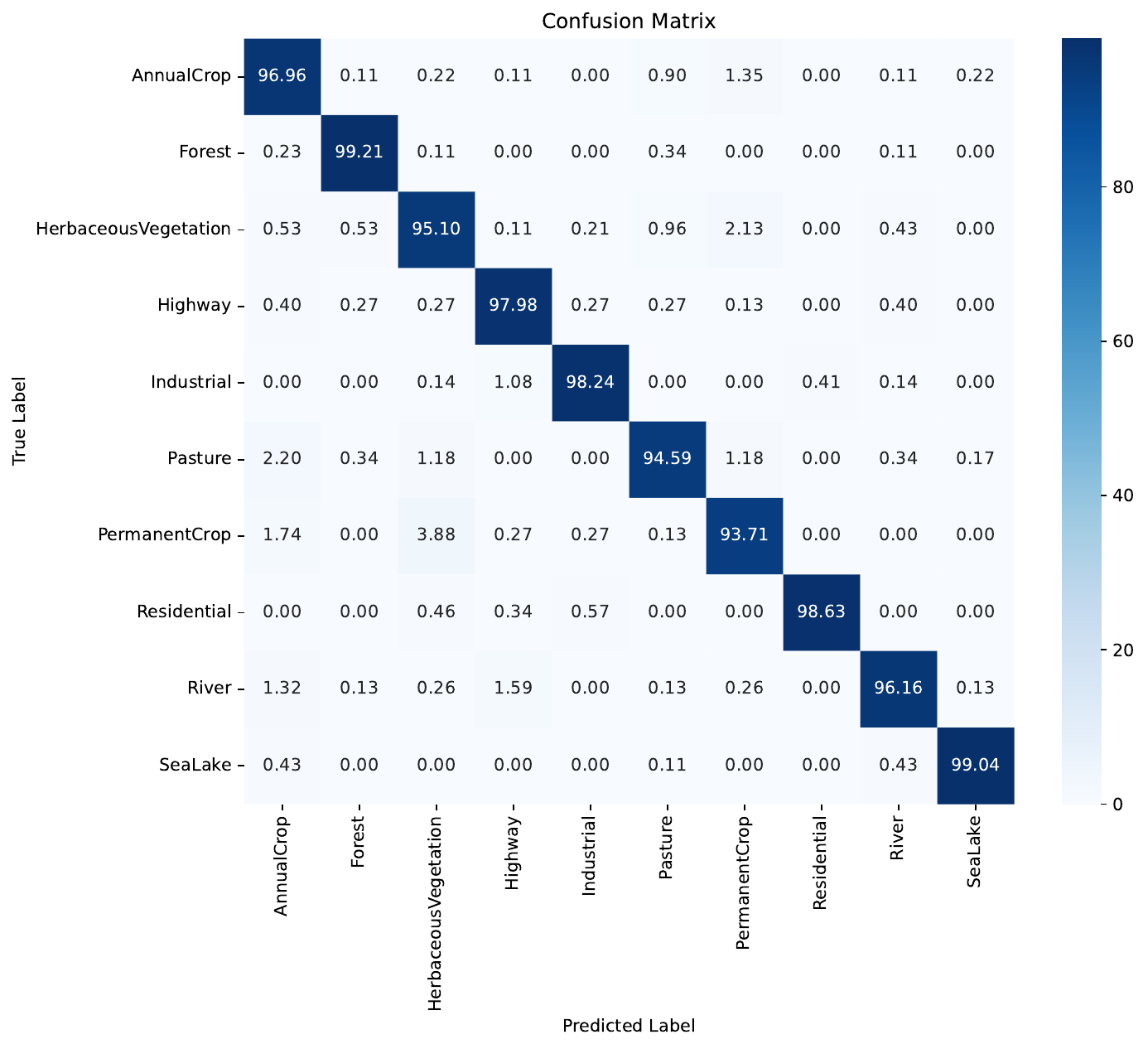}
    \includegraphics[scale=0.38]{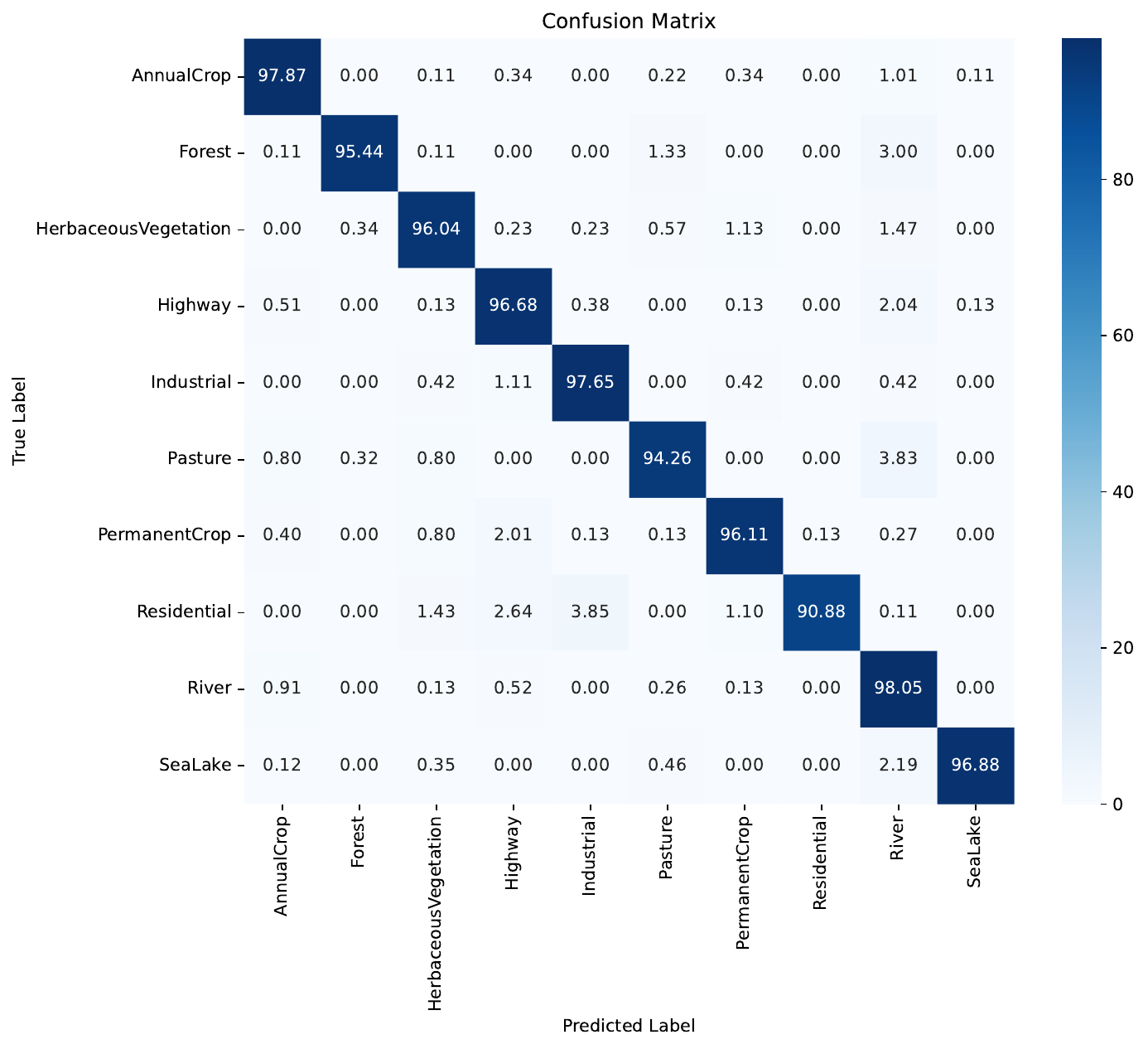}
    \caption{Top1 accuracy of confusion matrix using DT-JSCC based on 16PSK AWGN channel where PSNR=14dB (Upper K=128 and lower K=32).}
    \label{fig:cm_awgn}
\end{figure}

In comparing the confusion matrices of Fig.\ref{fig:cm_awgn}, we can observe some key performance differences over the AWGN channel where PSNR=14dB. For k=128, the classification accuracy across most categories is notably high, with certain categories like "Forest" showing a strong prediction rate of 99.21\%, while there are minor misclassifications in categories like "PermanentCrop" and "Pasture" with accuracies of 93.71\% and 94.59\% respectively. Meanwhile, the case of K=32 also performs well but demonstrates slightly different classification behaviors, such as a stronger prediction for "River" at 98.05\% but more substantial confusion in categories like "Forest" and "Residential," where there are larger misclassification rates with accuracies of 95.44\% and 90.88\% compared to the K=128 performance. The identical trend of K=32 and K=128 can be observed over the 16APSK Rician channel. 

\begin{figure}[!ht]
    \centering
    \includegraphics[scale=0.38]{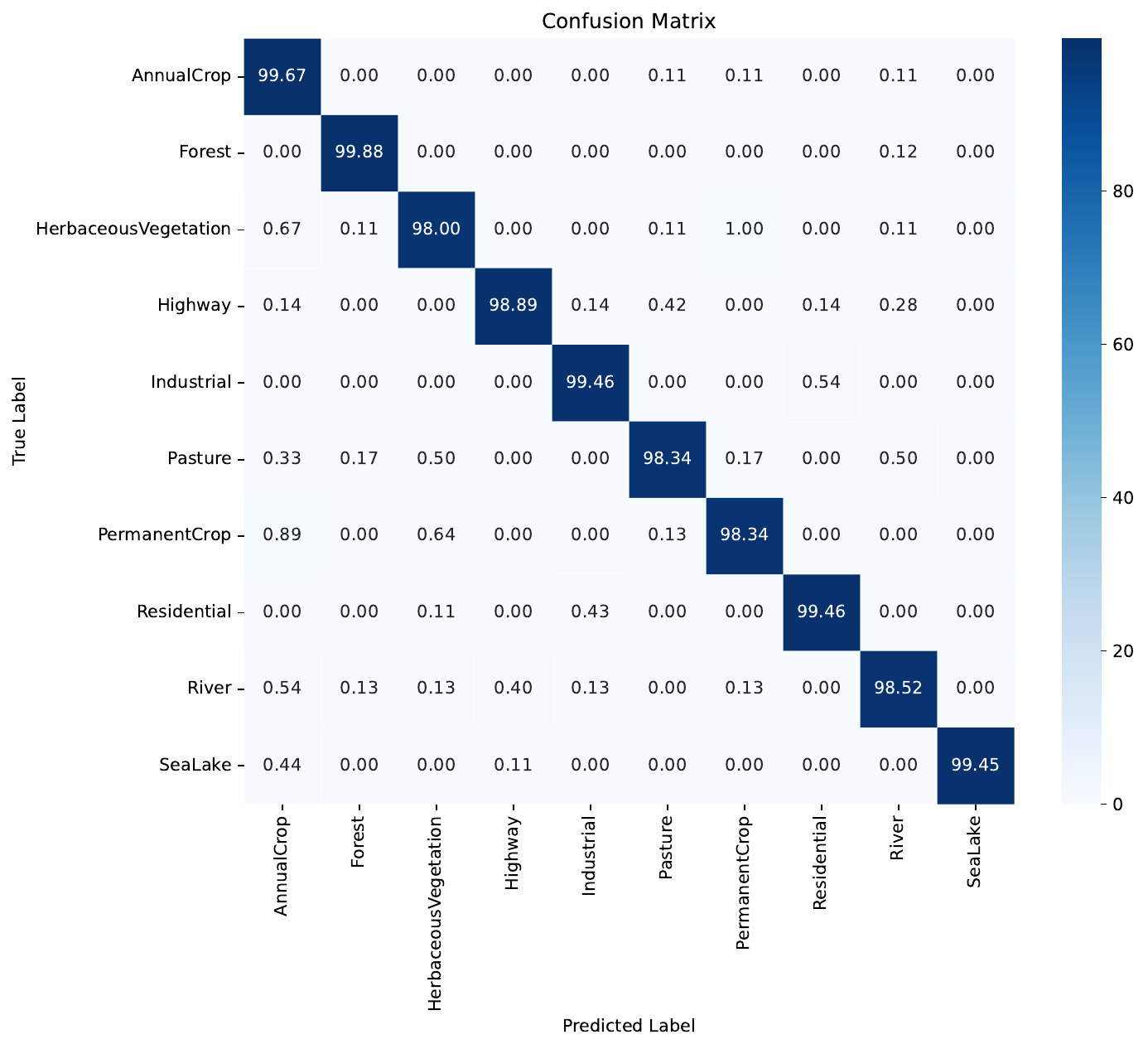}
    \caption{Top1 accuracy of confusion matrix using DTJSCC based on 16APSK Rician channel where PSNR=12dB and K=128.}
    \label{fig:cm_rician}
\end{figure}

As illustrated in the confusion matrices of Fig.\ref{fig:cm_rician}, we can observe some key performance differences for DT-JSCC using K=128, the classification accuracy across most categories is notably high, with certain categories like "Forest" showing a strong prediction rate of 99.88\%, while there are minor misclassifications in categories like "PermanentCrop" and "Pasture" with accuracies of 98.34\%.

\section{Conclusion}
In this paper, we explored the critical role of semantic inference models and deep learning in enhancing EO systems, focusing on the integration of data-driven modeling approaches. With the rise of big EO data and the increasing complexity of satellite imagery, conventional methods struggle to meet the demands of modern applications such as environmental monitoring, disaster management, and resource allocation. By utilizing advanced deep learning techniques, we have illustrated the potential to overcome key challenges, including the scarcity of labeled data, the variability in data sources, and the intricacies of multi-dimensional datasets. The incorporation of semantic inference enables systems to extract meaningful insights from vast and complex datasets, paving the way for more accurate decision-making in both civilian and critical operations. The advancements in on-board processing hardware demonstrate the increasing feasibility of efficient and adaptable computation for space missions. The evolution of SRAM-based FPGAs, radiation-hardened processors, and reconfigurable MPSoCs has opened new avenues for enhancing spacecraft resilience, adaptability, and task performance in real time. These systems ensure precise handling of computational tasks while mitigating the effects of space radiation, a critical capability for sustained mission success. Ultimately, on-board processing solutions promise improved data handling and communication efficiency, aligning with the goals of modern satellite technology and paving the way for sophisticated applications in space exploration.

%This work presents an advanced framework for semantic communication networks, with a focus on the development and optimization of cognitive semantic EO systems. By incorporating semantic-aware methodologies like JSCC and DT-JSCC, the proposed system enhances communication efficiency, particularly in satellite networks. The integration of semantic data augmentation, inter-satellite links, and cognitive semantic processing ensures that only the most relevant, task-specific data is transmitted, significantly reducing communication overhead and improving system performance. Furthermore, by leveraging semantic inference and cognitive augmentation techniques, the system facilitates better image analysis, change detection, and decision-making in real-time applications. The proposed end-to-end model demonstrates how integrating semantic cognition into satellite systems can drastically improve data transmission, interpretation, and overall system efficiency, thereby addressing the demands of 6G networks and beyond. This research lays the groundwork for future advancements in semantic communication, with promising applications in a wide range of fields, including remote sensing, autonomous systems, and smart communication networks.

%\section*{Acknowledgment}

\bibliographystyle{IEEEtran}
\bibliography{main}

\end{document}